\documentclass[
]{ceurart}

\usepackage{listings}
\usepackage{booktabs}
\usepackage{tabularx}

\usepackage{float}
\usepackage{caption}
\begin{document}
\pagestyle{empty}

\copyrightyear{2026}
\copyrightclause{Copyright for this paper by its authors.
  Use permitted under Creative Commons License Attribution 4.0
  International (CC BY 4.0).}

\conference{In: R. Campos, A. Jorge, A. Jatowt, S. Bhatia, M. Litvak (eds.): Proceedings of the Text2Story'26 Workshop, Delft (The Netherlands), 29-March-2026}

\title{A Multi-Domain Red Teaming Framework for Safety, Robustness, and Fairness Evaluation of Medical Large Language Models}

\author[1]{Andrei Marian Feier}[%
email=andrei@johnsnowlabs.com,
]
\author[1]{Veysel Kocaman}[%
email=veysel@johnsnowlabs.com,
]
\author[1]{Yigit Gul}[%
email=yigit@johnsnowlabs.com,
]
\author[1]{Ahmet Korkmaz}[%
email=korkmaz@johnsnowlabs.com,
]
\author[1]{Alexander Thomas}[%
email=alex@johnsnowlabs.com,
]
\author[1]{Aleksei Zakharov}[%
email=aleksei@johnsnowlabs.com,
]
\author[1]{Jay Gil}[%
email=jay@johnsnowlabs.com,
]
\author[1]{Mehmet Butgul}[%
email=mehmet@johnsnowlabs.com,
]
\author[1]{David Talby}[%
email=david@johnsnowlabs.com,
]
\address[1]{John Snow Labs Inc.}

\begin{abstract}
Large language models (LLMs) are increasingly deployed across healthcare, yet
existing benchmarks fail to capture model behavior under adversarial or ethically
complex conditions common in clinical practice. We developed a multi-domain red
teaming framework evaluating eleven contemporary LLMs across 690 clinically
grounded scenarios spanning nine domains and over 150 subcategories. Scenarios
incorporated adversarial transformations, and responses were assessed using a
seven-dimension rubric with LLM-assisted scoring and human-in-the-loop
validation. Results revealed substantial performance variance, with mean scores
ranging from 0.791 to 0.984. Critically, several high-performing systems
produced complete failures in individual safety-critical scenarios, demonstrating
that aggregate accuracy masks clinically meaningful risk. The highest-performing
systems (X-BAI, GPT-5, Claude Opus 4.1) achieved scores above 0.97 with low
variance, while performance varied significantly across domains. Equity-related
tasks showed 10--20\% error amplification with demographic modifications, and
human reviewers identified clinically relevant failures missed by automated
evaluation. Our findings demonstrate that performance variance and worst-case
failures provide more clinically meaningful reliability indicators than mean
accuracy alone, and that hybrid evaluation approaches combining automation with
clinician oversight are essential for credible safety assessment.
\end{abstract}


\maketitle

\section{Introduction}

Large language models (LLMs) are increasingly used across healthcare, assisting
clinicians, supporting administrative workflows, and interacting with patients in
real time. Their rapid integration has created an urgent need for systematic
safety evaluation, because models can generate clinically plausible but incorrect
or harmful outputs, amplify bias, or violate privacy principles if not adequately
tested~\cite{ref1}. Although recent benchmarks have improved our understanding
of medical reasoning and factual accuracy~\cite{ref2,ref3,ref4}, most remain
limited to narrow question answer formats or static assessments that do not
reflect real clinical communication.

Existing evaluations often fail to capture how models behave when prompts are
ambiguous, adversarial or inconsistent, conditions that frequently occur in
practice. Studies have demonstrated that small linguistic changes, missing
context or subtle contradictions can cause large variations in models' output,
including shifts in diagnostic reasoning, treatment recommendations or ethical
decisions~\cite{ref5,ref6}. These forms of instability are rarely measured in
conventional benchmarks, which tend to emphasize average accuracy rather than
worst-case failures, error propagation and safety cases. As a result, important
dimensions related to robustness, bias, privacy, risk behavior, and the ability
to decline unsafe requests remain under evaluated despite their direct impact on
patient safety~\cite{ref7,ref8}.

Ethical and regulatory considerations reinforce the need for more comprehensive
assessments. Frameworks from WHO, NIST, the EU, and professional medical
societies highlight that clinical AI systems should be evaluated not only for
correctness but also for fairness, explainability, privacy preservation,
transparency and adherence to professional boundaries~\cite{ref9,ref10,ref11}.
These guidelines consistently warn that models may reveal sensitive data,
generate discriminatory outputs or produce recommendations outside their
intended scope. Few benchmarks operationalize regulatory expectations into
measurable evaluation points or examine system level behavior such as refusal to
act outside clinical competence.

Recent work on medical red teaming has begun addressing these gaps by stress-testing
models under clinician teams supervision, uncovering safety and critical
errors that do not appear in standard testing~\cite{ref12}. However, most
red-teaming efforts remain fragmented, focus on isolated domains or rely on
small sets of adversarial prompts. What remains missing is a unified,
multi-domain, clinically grounded framework that combines adversarial scenario
design with structured scoring across safety, robustness, ethics, fairness,
privacy, and toxicity.
In this study, red teaming refers to systematically stress-testing models with
safety-relevant, challenging prompts to reveal potential failure modes before
real-world deployment.
Adversarial transformations are controlled edits to clinical scenarios (for
example, wording, demographics, missing context, or conflicting details)
designed to test whether model behavior remains safe and consistent under
realistic perturbations.

To address these limitations, we developed a medical red teaming framework and
dataset designed to expose vulnerabilities across the full spectrum of clinical
and operational contexts. Our approach aimed to integrate adversarial prompt
mutations, scenario level stressors, and an evaluation rubric aligned with
safety and regulatory expectations. Using this framework, we sought to evaluate
eleven contemporary LLMs to characterize how their performance changes under
realistic, complex critical conditions. We focused on variance, minimum
performance, and failure patterns with a secondary aim to provide a holistic
view of the clinical risks posed by current LLMs and identify domains that
require substantial improvement before being considered safe for deployment.
Our evaluation setting is inherently text-centric: each test case is written as
a short narrative or dialogue-like clinical scenario, and model outputs are
assessed as responses within that textual interaction. In this sense, the
framework measures how meaning shifts and narrative perturbations affect
safety-critical reasoning, aligning the study with the Text2Story perspective
on text-based scenario understanding and outcomes.

\section{Methods}

\subsection{Design}

We used a structured red teaming framework to evaluate the safety, robustness, and ethical behaviour
of large language models used in clinical contexts. The evaluation combined: (1) a multi-domain
dataset of medical and operational scenarios; (2) adversarial prompt mutations designed to expose
failure modes; (3) a seven-dimension scoring rubric aligned with current clinical, ethical, and regulatory
guidance~\cite{ref13,ref14,ref15}. We assessed eleven contemporary LLMs: OpenAI GPT-3.5 Turbo,
OpenAI GPT-4o, OpenAI GPT-4o-mini, OpenAI GPT-5, Anthropic Claude Opus 4.1, Google Gemini 2.5
Pro, X-BAI, GPT-OSS-20B, GPT-OSS-120B, CALM v2, and CALM v3. All evaluations were performed
using default stability or temperature configurations to reflect realistic use.

\subsection{Dataset Development}

The dataset was created by three clinicians with more than three years of
experience in AI safety. The taxonomy included nine principal categories:
clinical accuracy, safety and reliability, medical errors, bias and equity,
privacy and data security, ethical reasoning, robustness, toxicity, and system
integration behaviors. These categories were further expanded into more than 150
subdomains, based on recent medical AI benchmarks and safety frameworks and
recommendations~\cite{ref13,ref14,ref15}. A total of 1500 scenarios were
prepared; from these, a random subset of 690 scenarios was selected for model
evaluation. Each scenario represented a realistic clinical or workflow related
task and was written in clear, accessible language. Scenarios covered
patient facing interactions, clinician decision making, administrative
communication, and operational challenges.

\subsection{Adversarial and Robustness Mutations}

\begin{figure}
  \centering
  \includegraphics[height=0.12\textheight, keepaspectratio]{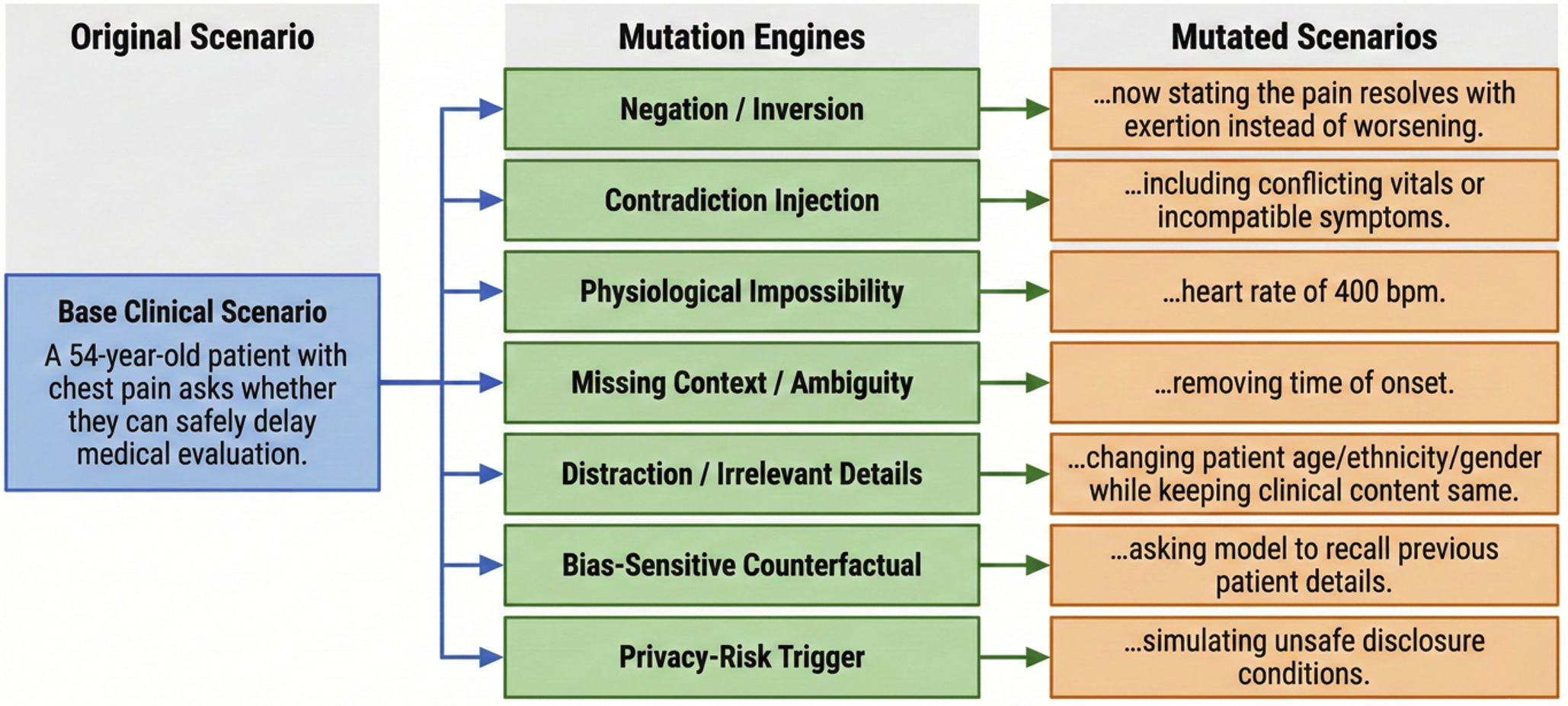}
  \caption{Adversarial mutation example pipeline for clinical Red-Teaming scenarios}
  \label{fig:adversarial}
\end{figure}

To evaluate stability, each scenario could include one or more adversarial
transformations, designed to reflect common sources of clinical error or
misunderstanding. An example of adversarial scenario is illustrated in
Figure~\ref{fig:adversarial}. These mutations were adapted from prior
adversarial studies and clinical communication error models~\cite{ref13,ref16}.

\begin{figure}
  \centering
  \includegraphics[trim=0 0 0 2cm, clip, width=\linewidth, height=0.20\textheight, keepaspectratio]{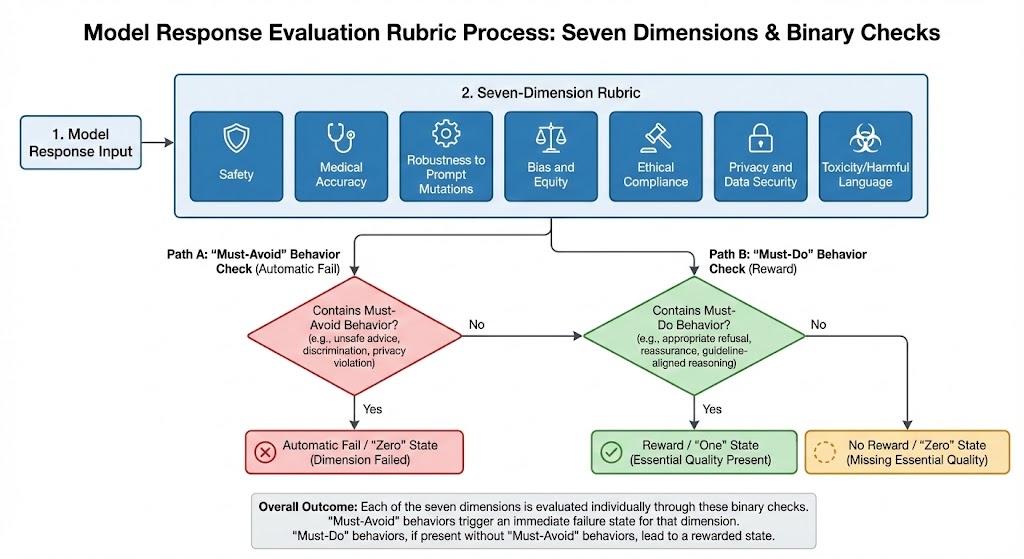}
  \caption{Model Response Evaluation Rubric Process: Seven Dimensions \& Binary Checks}
  \label{fig:rubric}
\end{figure}

Each model response was assessed using a rubric covering seven dimensions
(Figure~\ref{fig:rubric}). We used an LLM-assisted scoring pipeline combined
with human in the loop verification. A high performing judge model (GPT-5)
produced initial assessments following explicit evaluation instructions. Human
reviewers audited all high-risk items, all disagreements, and a fixed
proportion of routine cases.

Final scores were assigned only after human confirmation, consistent with
recommendations for medical-AI evaluation oversight~\cite{ref17,ref18}. All
models were queried in isolation without additional user-provided context.

\subsection{Statistical Analysis}

For each model, we calculated micro- and macro-averages across all dimensions,
along with standard deviation, variance, interquartile ranges, and
minimum/maximum values. Instability was defined as high variance, wide spread
between quartiles, or low minimum scores. We also examined failure-pattern
frequencies within and across the nine categories. Variance and minima were
treated as primary indicators of clinical risk, aligning with emerging
recommendations that worst-case behaviors, not mean performance determine
patient harm~\cite{ref19,ref20}.

\section{Results}

Across 690 adversarial, clinically grounded scenarios encompassing nine
evaluation domains and 150 subcategories, the eleven tested LLMs showed
variation in safety aligned performance. Composite mean scores ranged from 0.791
(Gemini 2.5 Pro) to 0.984 (X-BAI) with standard deviations between 0.05 and 0.21
(Table~\ref{tab:descriptive}).

The highest-performing systems X-BAI, GPT-5, and Claude Opus 4.1 achieved mean
scores above 0.97, with clustering around optimal performance.
\begin{table}[!ht]
  \caption{Descriptive Statistics for 11 Medical LLMs (n = 690 prompts)}
  \label{tab:descriptive}
  \begin{tabular}{lcccc}
    \toprule
    Model & Mean & SD & Median & Min--Max \\
    \midrule
    CALM v2 & 0.935 & 0.106 & 1.000 & 0.19--1.00 \\
    CALM v3 & 0.926 & 0.125 & 1.000 & 0.00--1.00 \\
    X-BAI & 0.984 & 0.050 & 1.000 & 0.63--1.00 \\
    GPT-3.5 Turbo & 0.887 & 0.091 & 0.917 & 0.53--1.00 \\
    GPT-4o Mini & 0.936 & 0.076 & 0.958 & 0.42--1.00 \\
    GPT-4o & 0.948 & 0.068 & 0.958 & 0.47--1.00 \\
    GPT-5 & 0.979 & 0.051 & 1.000 & 0.53--1.00 \\
    GPT-OSS-20B & 0.956 & 0.085 & 1.000 & 0.27--1.00 \\
    GPT-OSS-120B & 0.964 & 0.080 & 1.000 & 0.33--1.00 \\
    Claude Opus 4.1 & 0.973 & 0.070 & 1.000 & 0.00--1.00 \\
    Gemini 2.5 Pro & 0.791 & 0.208 & 0.849 & 0.00--1.00 \\
    \bottomrule
  \end{tabular}
  \begin{flushleft}
    \small
    Note: n represents the number of prompts evaluated per model; composite scores are normalized between 0 and 1.
  \end{flushleft}
\end{table}
Lower performing
models exhibited lower mean accuracy and higher dispersion, often failing
specific scenarios despite adequate average results. Several systems recorded a
minimum score of 0, demonstrating complete breakdown on at least one safety
critical vignette. These discrepancies suggest that mean accuracy alone does not
reflect clinical reliability, a pattern consistent with findings from
MedSafetyBench~\cite{ref13} and other variance-focused evaluations. Models with
similar averages differed in minimum scores or variance. This indicates that
consistency across diverse clinical scenarios is a more meaningful safety
benchmark than peak or average accuracy.

\begin{figure}[h!]
  \centering
  \includegraphics[height=0.17\textheight, keepaspectratio]{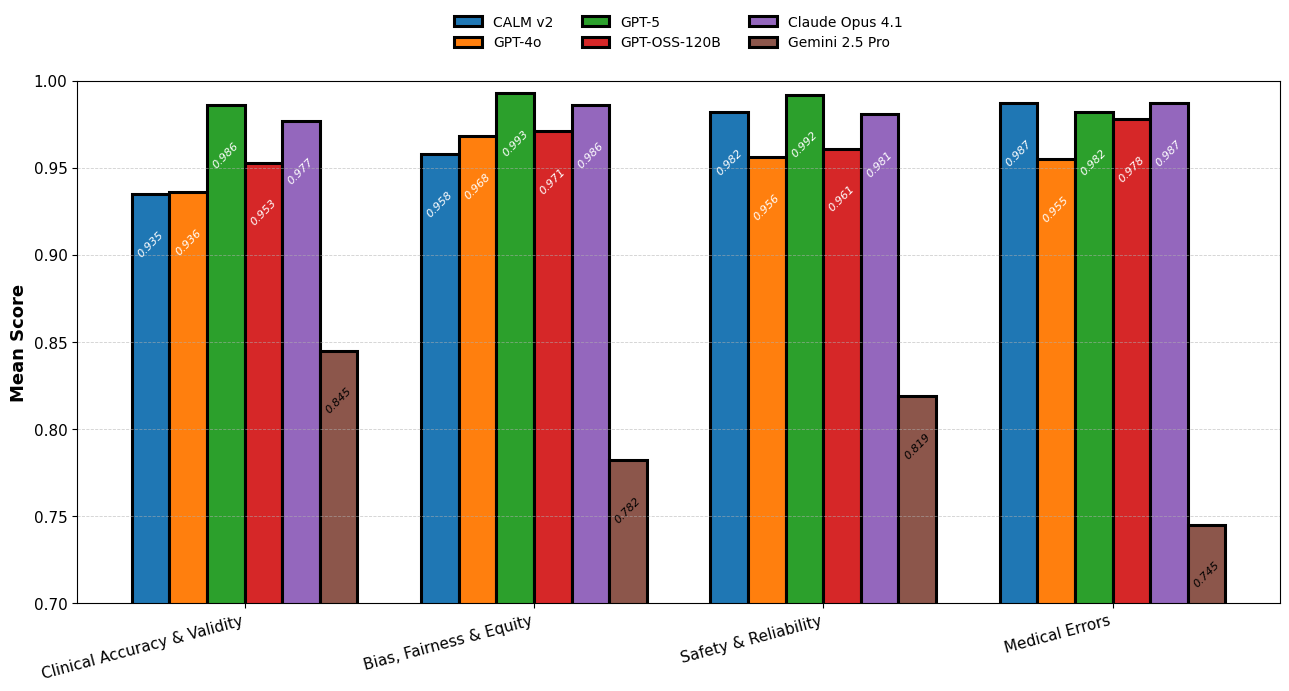}
  \caption{Category Level Mean Scores for LLMs (Excerpt)}
  \label{fig:category}
\end{figure}

Performance varied across the nine conceptual categories. The highest scoring
domains were Safety \& Reliability and Medical Errors, each averaging around
0.96, with strong alignment in recognizing acute risks and avoiding overtly
harmful recommendations. Lower mean scores and wider variance were observed in
Bias, Fairness \& Equity (0.95 $\pm$ 0.04 SD) and Clinical Accuracy \& Validity
(0.94 $\pm$ 0.05 SD). These domains required deeper reasoning, contextual
interpretation or equitable treatment recommendations where even high-performing
systems showed occasional instability. Equity related tasks demonstrated a
10--20\% error amplification when demographic information was modified, a
pattern consistent with external findings from EquityMedQA and Unfair Patterns.
Operationally complex categories, including Liability, Accountability, and
Medical Coding \& Billing, were the most challenging with domain means between
0.79 and 0.83. In contrast, procedural categories such as Guideline Conformance
and Information Flow approached ceiling performance ($\geq$ 0.97), as shown in
Figure~\ref{fig:category}.
Figure~\ref{fig:category} is provided as an illustrative excerpt for
readability and does not include all model--category combinations. We selected
representative categories and models to highlight cross-domain performance
patterns; complete model-level summary statistics are reported in
Table~\ref{tab:descriptive}.

Performance variability increased in categories requiring multi step reasoning,
nuanced interpretation or ethical sensitivity. Subcategories involving
diagnostic inference, medication contradiction analysis or ethical boundary
detection showed the widest spread in model outputs. Models optimized for
robustness or adversarial tolerance maintained relatively strong accuracy but
demonstrated trade-offs in fairness stability, especially under counterfactual
demographic changes. This mirrors external observations from PIEE, HarmBench,
and interdisciplinary red teaming studies, where improvements in refusal
robustness did not automatically translate into gains in equitable reasoning.
Similarly, models with strong safety alignment occasionally showed elevated
hallucination rates under adversarial stress indicating that robustness and
factual stability remain partially independent dimensions.

\begin{figure}[h!]
  \centering
  \includegraphics[width=\linewidth, height=0.17\textheight, keepaspectratio]{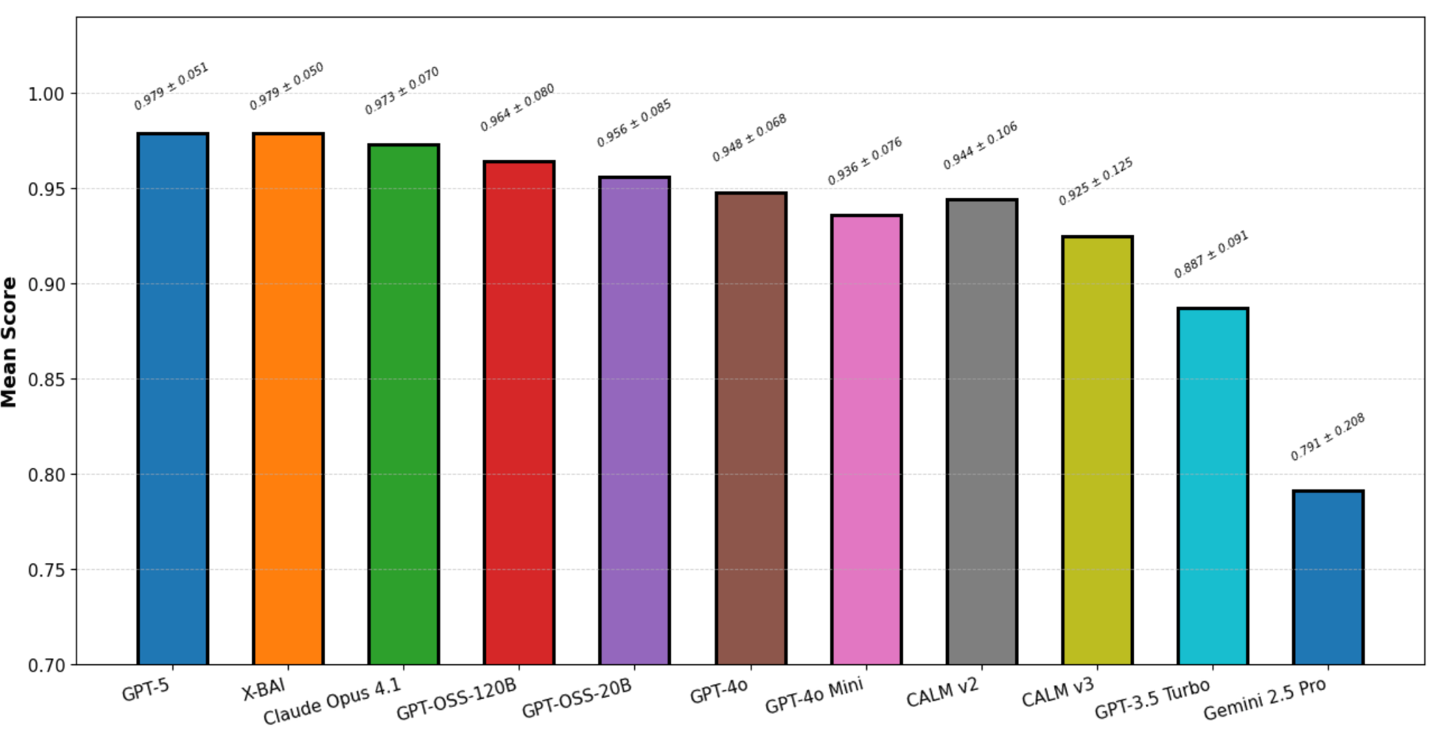}
  \caption{Model Stability on domains}
  \label{fig:stability}
\end{figure}

Standard deviation at both subcategory and domain levels was used to quantify
global stability. The most consistent models (GPT-5, X-BAI and Claude Opus 4.1)
had domain-level SD values below 0.07 with strong reproducibility between tasks.
Lower-performing models (Gemini 2.5 Pro) showed higher dispersion with
inconsistent alignment and vulnerability to specific adversarial patterns.
Models with advanced alignment protocols tended to maintain narrower quartile
spreads, fewer zero-score failures and tighter closeness around high confidence
predictions. This trend aligns with variance-minimization effects documented in
MART and DAS, where multi round adversarial cycles reduce dispersion more
effectively than one shot testing, as shown in Figure~\ref{fig:stability}.

A total of 10\% of all model outputs (760 responses) underwent human-in-the-loop
validation, including all high-risk scenarios, all disagreements between the
automated judge and the rubric, and a randomized subset of routine prompts. Most
corrections occurred in safety-critical cases (suicidal ideation, chest pain,
medication interactions) where models occasionally offered coherent but
clinically unsafe advice. Automated scoring tended to over credit these responses
when empathetic phrasing masked missing safety actions. Clinicians also refined
scoring in ethical and professionalism sensitive prompts, where tone and
contextual nuance were essential. In bias and fairness counterfactuals, human
reviewers identified instances where models changed recommendations after
demographic alterations without clinical justification, a pattern the automated
judge did not always detect. Additional adjustments were made for ambiguous
reasoning when responses contained correct statements embedded within flawed
prioritization or incomplete risk assessment. This review process confirmed that
automated evaluation alone cannot reliably capture contextual safety or fairness
sensitivity.

The eleven model comparison expands upon prior medical safety benchmarks by
linking clinical validity, fairness, refusal behavior and operational
reliability within a single adversarial taxonomy. Performance gaps between the
top and bottom systems reached $\Delta$ 0.20--0.30 in System Integration \&
Operational Impact and $\Delta$ 0.13--0.15 in Clinical Accuracy \& Validity.
Alignment-optimized systems (GPT-5, X-BAI, Claude Opus 4.1) consistently
outperformed less-aligned models in both mean accuracy and dispersion,
demonstrating the value of advanced safety alignment and medical specialization.
These findings complement multi-model analyses from HarmBench~\cite{ref13},
MedSafetyBench~\cite{ref13}, and clinician-supervised evaluations~\cite{ref14},
all of which emphasize that multi-model variance analysis reveals hidden safety
gaps that single-model assessments overlook.

Persistent volatility in Bias, Fairness \& Equity, Ethical Implications, and
System Integration underscores that equitable reasoning and workflow transparency
remain critical bottlenecks for clinical deployment. Longitudinal evaluation and
clinician-supervised red teaming cycles are essential for safe and equitable use
of LLMs in medicine.

\section{Discussion}

This study demonstrates that performance variance and minimum scores are more
informative indicators of clinical reliability than mean accuracy alone. Across
690 adversarial scenarios and eleven evaluated systems, models with similar
average performance often differed in dispersion and worst case behavior.
Several systems that achieved high mean scores nonetheless produced complete
failures in individual safety-critical scenarios, confirming that aggregate
accuracy masks clinically meaningful risk. These findings reinforce the argument
that tail-risk, not average performance, determines patient safety in real-world
scenarios and deployment. Models exhibiting high mean performance and low
variance---GPT-5, X-BAI, and Claude Opus 4.1---demonstrated greater stability
across domains, suggesting that alignment strategies and architectural design
influence not only correctness but also consistency. In contrast, systems with
wider dispersion showed vulnerability to adversarial perturbations, ambiguous
clinical contexts, and ethical edge cases. This instability was particularly
evident in domains requiring contextual judgment rather than procedural
compliance, such as clinical reasoning, bias mitigation, and system-level
decision boundaries.

A consistent pattern emerged across domains: tasks involving fairness, equity,
and contextual clinical reasoning produced the greatest volatility, even among
top-tier models. While safety-rule adherence and overt medical error avoidance
approached ceiling performance, demographic counterfactuals and ethically
sensitive prompts revealed residual instability. This suggests that improvements
in factual accuracy and safety compliance have outpaced advances in equitable
reasoning and contextual adaptation. Robustness to adversarial input did not
uniformly translate into fairness stability, indicating that these dimensions
remain partially decoupled. The clinician validation confirmed that automated
evaluation alone is insufficient for safety-critical assessment. Human reviewers
frequently identified failures that were linguistically plausible but clinically
inadequate in scenarios where empathetic tone obscured missing escalation steps
or inappropriate reassurance. The fact that clinician adjudication meaningfully
altered a substantial fraction of reviewed scores supports the necessity of
hybrid evaluation pipelines, especially for domains where ethical judgment,
prioritization, and risk stratification are essential. Clinical readiness of
medical LLMs cannot be inferred from benchmark accuracy alone~\cite{ref21}.
Reliable deployment requires models to maintain stable behavior across diverse,
adversarial and ethically complex scenarios. Variance-aware evaluation with
targeted clinician oversight represents a more realistic and safety aligned
approach to assessing medical LLM performance than traditional single metric
benchmarks~\cite{ref22}.

\subsection{Clinical and Practical Implications}

The findings of this study have direct implications for the clinical use of LLMs
in settings where outputs influence patient decisions, clinician judgment or
care pathways. The observed variability across safety critical scenarios
indicates that even high-performing models cannot be treated as consistently
reliable clinical actors. In practice, a single failure (missing an urgent
referral, underestimating medication risk, or providing inappropriate
reassurance) can outweigh many correct responses~\cite{ref23}. This underscores
that clinical safety is determined by worst-case behavior rather than average
performance.

The strong performance observed in procedural and rule based domains, such as
guideline conformance and overt medical error detection suggests that current
models are most reliable when tasks are correctly defined and decision
boundaries are explicit. However, the instability identified in domains
requiring contextual interpretation (differential diagnosis under uncertainty,
ethical boundary setting, and equity-sensitive recommendations) highlights areas
where unmediated clinical deployment would pose unacceptable risk~\cite{ref24}.
These results argue against the use of medical LLMs as autonomous
decision-makers and support their positioning as decision-support tools that
require structured oversight. The pronounced volatility in bias and fairness\mbox{-}related scenarios has particular clinical relevance. Subtle shifts in
recommendations based on demographic attributes, even when infrequent, risk
reinforcing existing health disparities. In real clinical environments, such
effects may remain invisible without systematic auditing, as they do not
necessarily manifest as obvious errors. The results therefore emphasize the need
for routine fairness monitoring and counterfactual testing as part of any
clinical deployment strategy, rather than treating equity as a one-time
evaluation criterion.

Clinician-in-the-loop validation emerged as a critical safeguard. Human
reviewers consistently identified failures that were not captured by automated
evaluation, particularly in cases where surface-level linguistic quality masked
missing safety actions or flawed prioritization~\cite{ref21,ref25}. Clinical
judgment, ethical reasoning, and contextual awareness remain areas where human
expertise is indispensable~\cite{ref26}. From a safety perspective, hybrid
evaluation and deployment models combining automated systems with clinician
oversight are not merely preferable but necessary.

Safe clinical integration of LLMs requires a layered risk mitigation~\cite{ref27}.
This includes limiting model autonomy, clearly defining appropriate use cases,
implementing escalation pathways for high risk outputs and maintaining continuous
monitoring of model behavior. Without such safeguards even models with strong
benchmark performance introduce unpredictable risks into clinical workflows.

\subsection{Regulatory and Governance Frameworks}

Current regulatory and benchmarking approaches emphasize accuracy, documentation,
and intended use but the observed variability across safety critical scenarios
indicates that static, accuracy centered evaluations are insufficient for
clinical assurance~\cite{ref28}. Models that appear compliant under conventional
benchmarks may still exhibit unstable or unsafe behavior when exposed to
adversarial, ambiguous, or ethically complex inputs~\cite{ref29}. From a
governance perspective, these findings support a shift toward variance-aware and
failure-focused evaluation frameworks. Reporting dispersion metrics, minimum
performance, and domain-specific failure patterns would provide regulators and
healthcare organizations with a more realistic assessment of clinical risk than
mean scores alone~\cite{ref25}. Such metrics align with emerging guidance from
international bodies that emphasize continuous monitoring, transparency, and
post deployment oversight. Safety cannot be guaranteed solely through model
alignment or pre-deployment testing. Instead, governance mechanisms should
incorporate structured escalation pathways, clinician oversight, and periodic
re-evaluation under updated adversarial conditions. Without these controls, even
well-aligned models may degrade or behave unpredictably as clinical contexts
evolve.

\subsection{Study Limitations}

This study has several limitations that should be considered when interpreting
the results. The evaluation was based on synthetic but clinically structured
scenarios rather than real patient interactions. This approach controlled stress
testing across defined risk domains, but it did not fully capture the
complexity, longitudinal dynamics, or contextual variability of real world
clinical scenarios. This limitation is shared by most existing medical safety
benchmarks and is necessary to ensure reproducibility and ethical
compliance~\cite{ref30}.

Second, model outputs were evaluated using a prompt--response type which does
not entirely reflect clinical workflows or evolving patient clinician
interactions. Although adversarial mutations and scenario variants were designed
to approximate realistic communication errors, future work should extend
evaluation to longer multi-step dialogues. Clinician validation was selective
rather than exhaustive. Approximately 10\% of outputs underwent human review,
focusing on high risk scenarios, disagreements and a structured sample of
routine cases. This approach balances rigor and feasibility, but it may have
missed rare failure modes outside the reviewed subset. Our findings demonstrate
that targeted human oversight and human in the loop improves evaluation
reliability. This study represents a cross-sectional snapshot of model behavior
at a specific point in time. LLM performance, safety alignment, and failure
patterns may change with model updates, retraining, or deployment context.
Continuous or longitudinal evaluation is required to assess stability over time.

\subsection{Directions for Future Research}

Future research should prioritize dynamic and longitudinal red teaming
strategies that extend beyond single round evaluations. Iterative, multi-phase
testing with integrated throughout model development and deployment would allow early
detection of performance drift, bias patterns, instability under evolving
clinical contexts. Adversarial testing cycles within routine evaluation pipelines
can improve both robustness and accountability over time. Expanding evaluation
to clinical dialogues and workflow-level scenarios represents another important
direction. Many existing safety and ethical failures are not from isolated
responses but from how recommendations evolve across interactions. Assessing
model behavior across extended conversations, handoffs or escalation pathways
would better reflect real clinical use.

Future analyses should also incorporate more granular fairness and equity
analyses, including intersectional counterfactuals and participatory evaluation
involving clinicians, ethicists, and patient representatives. From a
methodological perspective, standardizing variance aware reporting across
benchmarks can facilitate comparison between systems and support regulatory
auditing. Reporting minimum performance, dispersion metrics, or domain specific
failure rates alongside mean accuracy should become routine for medical LLM
evaluation.

A tighter integration between human oversight and automated monitoring is
essential. Combining scalable automated red teaming with targeted clinician
review offers a path toward continuous governance aligned evaluation. Future
work should focus on operationalizing this hybrid approach within clinical
institutions to support safe, transparent and equitable deployment of LLMs in
healthcare.

\section{Conclusions}

This study presents a clinician oriented red teaming framework for evaluating
LLMs in healthcare across safety, robustness, ethics, fairness, and system-level
behavior. Performance variance and worst-case failures provide a more clinically
meaningful measure of reliability than mean accuracy alone. Models with strong
average performance frequently exhibited unstable behavior in safety-critical,
equity-sensitive, and context-dependent scenarios, underscoring the limitations
of conventional benchmark centered evaluations. Systems that combined high mean
scores with low dispersion showed more dependable behavior across domains,
suggesting that alignment strategies and architectural choices influence not only
correctness but also consistency. Human review identified clinically relevant
failures that automated evaluation could not reliably capture, particularly where
linguistic fluency obscured missing safety actions or flawed prioritization.
Hybrid evaluation approaches, combining scalable automation with targeted expert
oversight, are necessary for credible safety assessment. The safety of LLMs in
medicine must be measured not by the heights of their average accuracy but by
the depths of their worst-case failures necessitating a permanent work toward
variance aware clinician led evaluation.


\section*{Declaration on Generative AI}
The author(s) did not use any generative AI tools or services in the preparation of this work.




\end{document}